%% file: paper.tex
\documentclass{article}

\usepackage[preprint]{configuration/neurips_2025}

\usepackage[utf8]{inputenc} % allow utf-8 input
\usepackage[T1]{fontenc}    % use 8-bit T1 fonts
\usepackage{hyperref}       % hyperlinks
\usepackage{url}            % simple URL typesetting
\usepackage{booktabs}       % professional-quality tables
\usepackage{amsfonts}       % blackboard math symbols
\usepackage{nicefrac}       % compact symbols for 1/2, etc.
\usepackage{microtype}      % microtypography
\usepackage{xcolor}         % colors

\input{resources/preamble.tex}

\title{Fast and Stable Diffusion Planning through Variational Adaptive Weighting}

\author{Zhiying Qiu$^{1}$ \quad Tao Lin$^{1,}\thanks{Corresponding author.} $ \\
$^{1}$Westlake University \\
{\tt\small qiuzhiying@westlake.edu.cn, lintao@westlake.edu.cn} \\
}

\begin{document}

\maketitle

\input{resources/main.tex}

\bibliography{resources/reference}

%%%%%%%%%%%%%%%%%%%%%%%%%%%%%%%%%%%%%%%%%%%%%%%%%%%%%%%%%%%%

\appendix

\input{resources/appendix.tex}
\end{document}

%% file: resources/preamble.tex
\usepackage{amsmath,amssymb,amsthm}

\providecommand{\tt}{\mathbf{t}}

\newenvironment{talign*}
{\csname align*\endcsname}
{\endalign}

\usepackage[utf8]{inputenc}         %
\usepackage[T1]{fontenc}            %
\usepackage{url}                    %
\usepackage{booktabs}               %
\usepackage{amsfonts}               %
\usepackage{nicefrac}               %
\usepackage{microtype}              %
\usepackage{xcolor}                 %
\usepackage{algorithm}
\usepackage{algorithmic}
\usepackage{graphicx}
\usepackage{subcaption}
\usepackage[flushleft]{threeparttable}
\usepackage{float}
\usepackage{multirow}
\usepackage{xspace}
\usepackage{natbib}
\usepackage{enumitem}
\usepackage[font=small]{caption}
\usepackage{autobreak}
\usepackage{sidecap}
\usepackage{wrapfig}
\usepackage{bbding}
\usepackage[toc, page, header]{appendix}
\usepackage{tikz}
\usepackage{xcolor}
\usepackage{pifont}
\usepackage{mdframed}
\usepackage{colortbl}

\hypersetup{
  colorlinks=true,
  linkcolor=blue,
  citecolor=blue,
  urlcolor=blue
}

\definecolor{coral}{RGB}{255,127,80}
\definecolor{darkgreen}{RGB}{0,100,0}
\definecolor{darkyellow}{RGB}{204,153,0}
\definecolor{salmon}{RGB}{250,128,114}

\newcommand{\transparentred}[1]{%
  \tikz[baseline=(X.base)] \node[fill=red, fill opacity=0.1, text opacity=1, inner sep=2pt, outer sep=0pt] (X) {#1};%
}
\newcommand{\thmref}[1]{\hyperref[#1]{\transparentred{Theorem~\ref*{#1}}}}

\newcommand{\transparentgray}[1]{%
  \tikz[baseline=(X.base)] \node[fill=gray, fill opacity=0.1, text opacity=1, inner sep=2pt, outer sep=0pt] (X) {#1};%
}
\newcommand{\defref}[1]{\hyperref[#1]{\transparentgray{Definition~\ref*{#1}}}}

\newcommand{\transparentblue}[1]{%
  \tikz[baseline=(X.base)] \node[fill=blue, fill opacity=0.1, text opacity=1, inner sep=2pt, outer sep=0pt] (X) {#1};%
}
\newcommand{\propref}[1]{\hyperref[#1]{\transparentblue{Proposition~\ref*{#1}}}}

\newcommand{\transparentgreen}[1]{%
  \tikz[baseline=(X.base)] \node[fill=green, fill opacity=0.1, text opacity=1, inner sep=2pt, outer sep=0pt] (X) {#1};%
}
\newcommand{\assumpref}[1]{\hyperref[#1]{\transparentgreen{Assumption~\ref*{#1}}}}

\newcommand{\transparentyellow}[1]{%
  \tikz[baseline=(X.base)] \node[fill=yellow, fill opacity=0.1, text opacity=1, inner sep=2pt, outer sep=0pt] (X) {#1};%
}
\newcommand{\remarkref}[1]{\hyperref[#1]{\transparentyellow{Remark~\ref*{#1}}}}

\newcommand{\transparentpurple}[1]{%
  \tikz[baseline=(X.base)] \node[fill=purple, fill opacity=0.1, text opacity=1, inner sep=2pt, outer sep=0pt] (X) {#1};%
}
\newcommand{\hypref}[1]{\hyperref[#1]{\transparentpurple{Hypothesis~\ref*{#1}}}}

\newcommand{\transparentorange}[1]{%
  \tikz[baseline=(X.base)] \node[fill=orange, fill opacity=0.1, text opacity=1, inner sep=2pt, outer sep=0pt] (X) {#1};%
}
\newcommand{\conjref}[1]{\hyperref[#1]{\transparentorange{Conjecture~\ref*{#1}}}}

\newcommand{\transparentcyan}[1]{%
  \tikz[baseline=(X.base)] \node[fill=cyan, fill opacity=0.1, text opacity=1, inner sep=2pt, outer sep=0pt] (X) {#1};%
}
\newcommand{\lemref}[1]{\hyperref[#1]{\transparentcyan{Lemma~\ref*{#1}}}}

\newcommand{\transparentmagenta}[1]{%
  \tikz[baseline=(X.base)] \node[fill=magenta, fill opacity=0.1, text opacity=1, inner sep=2pt, outer sep=0pt] (X) {#1};%
}
\newcommand{\corref}[1]{\hyperref[#1]{\transparentmagenta{Corollary~\ref*{#1}}}}

\newcommand{\transparentpink}[1]{%
  \tikz[baseline=(X.base)] \node[fill=pink, fill opacity=0.1, text opacity=1, inner sep=2pt, outer sep=0pt] (X) {#1};%
}
\newcommand{\noteref}[1]{\hyperref[#1]{\transparentpink{Notation~\ref*{#1}}}}

\newcommand{\transparentviolet}[1]{%
  \tikz[baseline=(X.base)] \node[fill=violet, fill opacity=0.1, text opacity=1, inner sep=2pt, outer sep=0pt] (X) {#1};%
}
\newcommand{\claimref}[1]{\hyperref[#1]{\transparentviolet{Claim~\ref*{#1}}}}

\newcommand{\transparentsalmon}[1]{%
  \tikz[baseline=(X.base)] \node[fill=salmon, fill opacity=0.1, text opacity=1, inner sep=2pt, outer sep=0pt] (X) {#1};%
}
\newcommand{\probref}[1]{\hyperref[#1]{\transparentsalmon{Problem~\ref*{#1}}}}

\newcommand{\transparentlavender}[1]{%
  \tikz[baseline=(X.base)] \node[fill=lavender, fill opacity=0.1, text opacity=1, inner sep=2pt, outer sep=0pt] (X) {#1};%
}
\newcommand{\obsref}[1]{\hyperref[#1]{\transparentlavender{Observation~\ref*{#1}}}}

\newcommand{\transparentlime}[1]{%
  \tikz[baseline=(X.base)] \node[fill=lime, fill opacity=0.1, text opacity=1, inner sep=2pt, outer sep=0pt] (X) {#1};%
}
\newcommand{\algref}[1]{\hyperref[#1]{\transparentlime{Algorithm~\ref*{#1}}}}

\newcommand{\transparentteal}[1]{%
  \tikz[baseline=(X.base)] \node[fill=teal, fill opacity=0.1, text opacity=1, inner sep=2pt, outer sep=0pt] (X) {#1};%
}
\newcommand{\figref}[1]{\hyperref[#1]{\transparentteal{Figure~\ref*{#1}}}}

\newcommand{\transparentdarkgreen}[1]{%
  \tikz[baseline=(X.base)] \node[fill=darkgreen, fill opacity=0.1, text opacity=1, inner sep=2pt, outer sep=0pt] (X) {#1};%
}
\newcommand{\tabref}[1]{\hyperref[#1]{\transparentdarkgreen{Table~\ref*{#1}}}}

\newcommand{\transparentdarkyellow}[1]{%
  \tikz[baseline=(X.base)] \node[fill=darkyellow, fill opacity=0.1, text opacity=1, inner sep=2pt, outer sep=0pt] (X) {#1};%
}
\newcommand{\secref}[1]{\hyperref[#1]{\transparentdarkyellow{Section~\ref*{#1}}}}

\newcommand{\transparentcoral}[1]{%
  \tikz[baseline=(X.base)] \node[fill=coral, fill opacity=0.1, text opacity=1, inner sep=2pt, outer sep=0pt] (X) {#1};%
}
\newcommand{\appref}[1]{\hyperref[#1]{\transparentcoral{Appendix~\ref*{#1}}}}

\newtheoremstyle{custom}
{1pt} % Space above
{1pt} % Space below
{\itshape} % Body font
{} % Indent amount
{\bfseries} % Theorem head font
{} % Punctuation after theorem head
{ } % Space after theorem head
{\thmname{#1} \thmnumber{#2} \thmnote{(#3)} . } % Theorem head spec

\theoremstyle{custom}

\newtheorem{innerdefinition}{Definition}
\newtheorem{innerproposition}{Proposition}
\newtheorem{innerassumption}{Assumption}
\newtheorem{innerremark}{Remark}
\newtheorem{innertheorem}{Theorem}
\newtheorem{innerhypothesis}{Hypothesis}
\newtheorem{innerconjecture}{Conjecture}
\newtheorem{innerlemma}{Lemma}
\newtheorem{innercorollary}{Corollary}
\newtheorem{innernotation}{Notation}
\newtheorem{innerclaim}{Claim}
\newtheorem{innerproblem}{Problem}

\newtheorem{innerobservation}{Observation}

\newmdenv[
  backgroundcolor=gray!10,
  linecolor=gray!100,
  linewidth=0.8pt,
  skipabove=2pt,
  skipbelow=2pt,
  innertopmargin=10pt,
  innerbottommargin=5pt,
  innerleftmargin=5pt,
  innerrightmargin=5pt,
]{definitionframe}

\newmdenv[
  backgroundcolor=blue!10,
  linecolor=blue!100,
  linewidth=0.8pt,
  skipabove=2pt,
  skipbelow=2pt,
  innertopmargin=10pt,
  innerbottommargin=5pt,
  innerleftmargin=5pt,
  innerrightmargin=5pt,
]{propositionframe}

\newmdenv[
  backgroundcolor=green!10,
  linecolor=green!100,
  linewidth=0.8pt,
  skipabove=2pt,
  skipbelow=2pt,
  innertopmargin=10pt,
  innerbottommargin=5pt,
  innerleftmargin=5pt,
  innerrightmargin=5pt,
]{assumptionframe}

\newmdenv[
  backgroundcolor=yellow!10,
  linecolor=yellow!100,
  linewidth=0.8pt,
  skipabove=2pt,
  skipbelow=2pt,
  innertopmargin=10pt,
  innerbottommargin=5pt,
  innerleftmargin=5pt,
  innerrightmargin=5pt,
]{remarkframe}

\newmdenv[
  backgroundcolor=red!10,
  linecolor=red!100,
  linewidth=0.8pt,
  skipabove=2pt,
  skipbelow=2pt,
  innertopmargin=10pt,
  innerbottommargin=5pt,
  innerleftmargin=5pt,
  innerrightmargin=5pt,
]{theoremframe}

\newmdenv[
  backgroundcolor=purple!10,
  linecolor=purple!100,
  linewidth=0.8pt,
  skipabove=2pt,
  skipbelow=2pt,
  innertopmargin=10pt,
  innerbottommargin=5pt,
  innerleftmargin=5pt,
  innerrightmargin=5pt,
]{hypothesisframe}

\newmdenv[
  backgroundcolor=orange!10,
  linecolor=orange!100,
  linewidth=0.8pt,
  skipabove=2pt,
  skipbelow=2pt,
  innertopmargin=10pt,
  innerbottommargin=5pt,
  innerleftmargin=5pt,
  innerrightmargin=5pt,
]{conjectureframe}

\newmdenv[
  backgroundcolor=cyan!10,
  linecolor=cyan!100,
  linewidth=0.8pt,
  skipabove=2pt,
  skipbelow=2pt,
  innertopmargin=10pt,
  innerbottommargin=5pt,
  innerleftmargin=5pt,
  innerrightmargin=5pt,
]{lemmaframe}

\newmdenv[
  backgroundcolor=magenta!10,
  linecolor=magenta!100,
  linewidth=0.8pt,
  skipabove=2pt,
  skipbelow=2pt,
  innertopmargin=10pt,
  innerbottommargin=5pt,
  innerleftmargin=5pt,
  innerrightmargin=5pt,
]{corollaryframe}

\newmdenv[
  backgroundcolor=pink!10,
  linecolor=pink!100,
  linewidth=0.8pt,
  skipabove=2pt,
  skipbelow=2pt,
  innertopmargin=10pt,
  innerbottommargin=5pt,
  innerleftmargin=5pt,
  innerrightmargin=5pt,
]{notationframe}

\newmdenv[
  backgroundcolor=violet!10,
  linecolor=violet!100,
  linewidth=0.8pt,
  skipabove=2pt,
  skipbelow=2pt,
  innertopmargin=10pt,
  innerbottommargin=5pt,
  innerleftmargin=5pt,
  innerrightmargin=5pt,
]{claimframe}

\newmdenv[
  backgroundcolor=salmon!10,
  linecolor=salmon!100,
  linewidth=0.8pt,
  skipabove=2pt,
  skipbelow=2pt,
  innertopmargin=10pt,
  innerbottommargin=5pt,
  innerleftmargin=5pt,
  innerrightmargin=5pt,
]{problemframe}

\newmdenv[
  backgroundcolor=lavender!10,
  linecolor=lavender!100,
  linewidth=0.8pt,
  skipabove=2pt,
  skipbelow=2pt,
  innertopmargin=10pt,
  innerbottommargin=5pt,
  innerleftmargin=5pt,
  innerrightmargin=5pt,
]{observationframe}

\usepackage[textwidth=2.cm,textsize=tiny]{todonotes}

\definecolor{darkgreen}{RGB}{0, 100, 0}

%% file: resources/main.tex
\begin{abstract}
Diffusion models have recently shown promise in offline RL. However, these methods often suffer from high training costs and slow convergence, particularly when using transformer-based denoising backbones. While several optimization strategies have been proposed -- such as modified noise schedules, auxiliary prediction targets, and adaptive loss weighting -- challenges remain in achieving stable and efficient training. In particular, existing loss weighting functions typically rely on neural network approximators, which can be ineffective in early training phases due to limited generalization capacity of MLPs when exposed to sparse feedback in the early training stages. In this work, we derive a variationally optimal uncertainty-aware weighting function and introduce a closed-form polynomial approximation method for its online estimation under the flow-based generative modeling framework. We integrate our method into a diffusion planning pipeline and evaluate it on standard offline RL benchmarks. Experimental results on Maze2D and Kitchen tasks show that our method achieves competitive performance with up to 10 times fewer training steps, highlighting its practical effectiveness.
\end{abstract}

\section{Introduction}

Decision-making based on offline data has become a foundational paradigm in robotics and artificial intelligence~\citep{bellman1957markovian}. This approach enables agents to acquire complex behaviors by learning from expert demonstrations, thereby avoiding the need for explicit programming or costly online exploration. Despite its practical advantages, offline learning presents fundamental challenges, particularly in long-horizon planning scenarios and environments with high-dimensional action spaces. In such settings, modeling long-term sequential dependencies and capturing the underlying distribution of expert behavior remains difficult for conventional methods~\citep{deisenroth2011pilco, parmas2018pipps}.

To address these issues, recent research has explored the use of diffusion models—originally developed for high-fidelity image and video generation~\citep{ho2020denoising, dhariwal2021diffusion}—as a generative approach for sequential decision-making. Diffusion models are capable of modeling complex distributions and long-range dependencies, making them well-suited for planning in continuous control tasks. Inspired by their success in generative modeling, several works have adapted diffusion models to trajectory planning settings~\citep{janner2022planning, ajay2022conditional, lu2023contrastive, li2023hierarchical}, where the goal is to generate plausible sequences of states and actions conditioned on initial observations or task objectives. Leveraging conditional sampling techniques such as classifier-free guidance~\citep{ho2020denoising, ho2021classifier}, these models can generate trajectories with desirable properties, such as achieving high rewards, thereby serving as effective planners in offline reinforcement learning~\citep{levine2020offline}.

Despite the notable progress of diffusion-based planners in offline reinforcement learning, a key limitation remains: the high training cost and slow convergence associated with diffusion models, a challenge particularly evident in Transformer-based denoising backbones. While these backbones achieve notable success, often outperforming U-Net, they still contend with slow convergence rates~\cite{dong2023aligndiff, lu2025makes}. To mitigate this issue, a number of optimization strategies have been explored, including alternative noise schedules~\cite{nichol2021improved, kingma2023understanding}, additional prediction targets such as velocity direction~\cite{yao2024fasterdit}, and loss weighting functions~\cite{esser2024scaling, hang2023efficient, karras2024analyzing}. Among these, loss weighting functions have shown promise in accelerating convergence and improving sample quality.

However, existing popular adaptive weighting function ~\cite{karras2024analyzing} is implemented via neural network approximators,  which are jointly optimized with the main model during training. While theoretically flexible, these approximators tend to exhibit poor empirical behavior in early training, when the per-noise training loss is noisy and highly dynamic. As a result, the learned weighting functions may degenerate to nearly constant outputs, failing to accurately reflect uncertainty across noise scales and leading to unstable or inefficient training dynamics. These observations motivate the need for a principled and stable loss weighting strategy that avoids the pitfalls of iterative neural approximation.

In this work, we propose an efficient and theoretically grounded weighting functions under the flow-based generative modeling framework. Building on the continuous noise conditioning formulation enabled by flow matching~\cite{lipman2022flow}, we derive the variationally optimal form of an uncertainty-aware weighting function for training diffusion models. This yields a closed-form target function $u^*(\sigma)$, which captures the log-loss landscape over the noise scale $\sigma$ and provides precise control over the per-noise contribution to the total training objective. To instantiate this function without introducing auxiliary networks, we further propose a streaming polynomial regression scheme that directly estimates $u^*(\sigma)$ online via lightweight least-squares fitting. This approach enables fast convergence, improved generalization across noise scales, and stable training dynamics throughout the optimization process.

Our method can be seamlessly integrated into existing diffusion planning pipelines. We demonstrate its effectiveness by incorporating it into a flow-based planner and evaluating performance on standard offline RL benchmarks. Experimental results across a diverse suite of Maze2D and Kitchen tasks show that our approach achieves state-of-the-art performance with significantly fewer training steps—often reducing training time by an order of magnitude—while preserving or improving final reward performance. These results highlight the practical value of principled loss weighting in advancing efficient diffusion-based planning.

\section{Preliminaries} \label{sec:preliminaries}
\subsection{Diffusion Models for Planning} \label{sec:dm4planning}

Diffusion models, particularly score-based generative models \cite{song2020score}, have recently gained traction in sequential decision-making. Pioneering frameworks such as Diffuser \cite{janner2022planning} and Decision Diffuser (DD) \cite{ajay2022conditional} demonstrated their potential for trajectory generation and planning. These efforts have since spurred a growing body of work expanding diffusion planning to diverse domains \cite{liang2023adaptdiffuser,du2023learning,chen2024simple,li2023hierarchical}.

\paragraph{Modeling trajectories as state-action sequences.}
One of the earliest diffusion-based planning frameworks is Diffuser \cite{janner2022planning}, which models entire trajectories as sequences of state-action pairs. Specifically, it treats planning as a conditional generation problem over trajectories, where each trajectory $\tau$ consists of a fixed-horizon rollout of states and actions:
\begin{equation}
    \tau = \begin{bmatrix} s_t, s_{t+1}, \dots, s_{t+H-1} \\ a_t, a_{t+1}, \dots, a_{t+H-1} \end{bmatrix}.
    \label{eq:diffuser_trajectory}
\end{equation}
A standard score-based diffusion model is then trained to denoise and recover such trajectories, conditioning on observations like initial state. This approach has since been extended in subsequent works \cite{liang2023adaptdiffuser,chen2024simple,li2023hierarchical}, illustrating the general applicability of modeling dense state-action sequences.

\paragraph{State-only planning with inverse dynamics.}
Later methods, such as Decision Diffuser (DD) \cite{ajay2022conditional}, propose a more modular approach by decoupling action prediction from trajectory generation. Instead of modeling state-action pairs, these approaches\cite{ajay2022conditional, du2023learning} generate only the sequence of states:
\begin{equation}
    \tau = [s_t, s_{t+1}, \dots, s_{t+H-1}],
    \label{eq:decision_diffuser_trajectory}
\end{equation}
and employ an inverse dynamics model to recover the corresponding actions. Given two consecutive states $(s_t, s_{t+1})$, the agent predicts the action $a_t$ that likely caused this transition: $a_t := f_\phi(s_t, s_{t+1})$.

\paragraph{Jump-step planning with sparse trajectories}
Building on this idea, DV \cite{lu2025makes} introduces jump-step planning, where the agent plans over temporally sparse trajectories. Instead of modeling consecutive state transitions, DV generates trajectories using a fixed stride $m$:
\begin{equation}
    \tau = [s_t, s_{t+m}, s_{t+2m}, \dots, s_{t+(H-1)m}],
    \label{eq:dv_trajectory}
\end{equation}
and similarly infers actions through an inverse dynamics model: $a_t := f_\phi(s_t, s_{t+m})$. Empirical studies in DV demonstrate that jump-step planning not only reduces the burden of modeling long sequences, but the combination with inverse dynamics often leads to better sample efficiency and planning performance compared to dense-step baselines in complex environments with high-dimensional action spaces.

\subsection{Flow Matching} \label{sec:flow_matching}

Flow Matching provides an ODE-based framework for generative modeling by defining a continuous path between a prior distribution $p(\mathbf{z})$ and a data distribution $p(\mathbf{x})$, and learning a time-dependent velocity field to traverse it.
Concretely, let $\mathbf{x} \sim p(\mathbf{x})$ be a data sample, and let $\mathbf{z} \sim p(\mathbf{z}) = \mathcal{N}(\mathbf{0}, \mathbf{I})$ be a noise sample. We define a deterministic interpolation $\mathbf{x}_t$ between $\mathbf{x}$ and $\mathbf{z}$ over continuous time $t \in [0, 1]$,
$
    \mathbf{x}_t = \gamma(t)\, \mathbf{x} + \alpha(t)\, \mathbf{z},
$
such that $\mathbf{x}_t$ traces a probability path from data to noise. The instantaneous velocity of this path is given by
$
    \frac{d \mathbf{x}_t}{dt} = \dot{\gamma}(t)\, \mathbf{x} + \dot{\alpha}(t)\, \mathbf{z}.
$
A neural network $\boldsymbol{F}_\theta(\mathbf{x}_t, t)$ is trained to approximate this time-dependent velocity field. Once trained, the model enables sampling by solving the reverse-time ordinary differential equation (ODE):
$
    \frac{d \mathbf{x}_t}{dt} = \boldsymbol{F}_\theta(\mathbf{x}_t, t),
$
from t = 1 to t = 0, starting from $\mathbf{z} \sim \mathcal{N}(\mathbf{0}, \mathbf{I})$.

The training objective is to regress the predicted velocity $\boldsymbol{F}_\theta(\mathbf{x}_t, t)$ to the ground-truth velocity of the interpolation path. So it is given by
\begin{equation}
    \mathcal{L}_{\text{FM}} = \mathbb{E}_{\mathbf{z}, \mathbf{x}, t} \left[ w(t) \left\| \boldsymbol{F}_\theta(\mathbf{x}_t, t) - \left( \dot{\gamma}(t)\, \mathbf{x} + \dot{\alpha}(t)\, \mathbf{z} \right) \right\|_2^2 \right]
\end{equation}\label{eq:unify_flow}
where $w(t)$ is a time-dependent weighting function.

\paragraph{Linear interpolant~\citep{lipman2022flow, liu2022flow}.}
One of the simplest instantiations of flow matching employs a linear interpolant path by setting $\gamma(t) = 1 - t$ and $\alpha(t) = t$, yielding the path $\mathbf{x}_t = (1 - t)\, \mathbf{x} + t\, \mathbf{z}$ for $t \in [0, 1]$, where $\mathbf{x} \sim p(\mathbf{x})$ and $\mathbf{z} \sim \mathcal{N}(\mathbf{0}, \mathbf{I})$. The instantaneous velocity along this path is constant and given by $\frac{d \mathbf{x}_t}{dt} = \mathbf{z} - \mathbf{x}$. Under the general flow matching framework, the reverse-time ODE defined by $\frac{d \mathbf{x}_t}{dt} = \boldsymbol{F}_\theta(\mathbf{x}_t, t)$ is integrated from $t = 1$ to $t = 0$, starting from $\mathbf{z} \sim \mathcal{N}(\mathbf{0}, \mathbf{I})$. The corresponding training objective becomes:
\begin{equation}
    \label{eq:linear_flow}
    \mathcal{L}_{\text{linear}} = \mathbb{E}_{\mathbf{x}, \mathbf{z}, t} \left[ w(t)\, \left\| \boldsymbol{F}_\theta(\mathbf{x}_t, t) - (\mathbf{z} - \mathbf{x}) \right\|_2^2 \right],
\end{equation}
where $w(t)$ is a time-dependent weighting function.

\paragraph{Trigonometric interpolant~\citep{lu2024simplifying}.}
This instance adopts a trigonometric interpolation path, which aligns with recent formulations that emphasize improved smoothness and variance control.
By setting $\gamma(t) = \cos(t)$ and $\alpha(t) = \sin(t)$, the interpolation path is defined as $\mathbf{x}_t = \cos(t)\, \mathbf{x} + \sin(t)\, \mathbf{z}$ for $t \in [0, \frac{\pi}{2}]$, where $\mathbf{x} \sim p(\mathbf{x})$ and $\mathbf{z} \sim \mathcal{N}(\mathbf{0}, \sigma_d^2\, \mathbf{I})$, with $\sigma_d$ being the standard deviation of the data distribution $p(\mathbf{x})$.
The velocity field is given by $\frac{d \mathbf{x}_t}{dt} = -\sin(t)\, \mathbf{x} + \cos(t)\, \mathbf{z}$.
Sampling proceeds by solving the reverse-time ODE $\frac{d \mathbf{x}_t}{dt} = \sigma_d \boldsymbol{F}_\theta(\frac{\mathbf{x}_t}{\sigma_d}, c_{\mathrm{noise}}(t))$ from $t = \frac{\pi}{2}$ to $t = 0$, where $c_{\mathrm{noise}}(t)$ denotes a transformation of $t$ used for time conditioning. The corresponding training loss is formulated as:
\begin{equation}
    \label{eq:trig_flow}
    \mathcal{L}_{\text{trig}} = \mathbb{E}_{\mathbf{x}, \mathbf{z}, t} \left[ w(t)\, \left\| \boldsymbol{F}_\theta(\frac{\mathbf{x}_t}{\sigma_d}, c_{\mathrm{noise}}(t)) - (-\sin(t)\, \mathbf{x} + \cos(t)\, \mathbf{z}) \right\|_2^2 \right],
\end{equation}
where $w(t)$ is a time-dependent weighting function.

\section{Related Work}
Recent works have explored replacing U-Net with Transformer-based denoising networks in diffusion models, particularly in diffusion planning tasks on offline RL benchmarks such as D4RL. Dong et al. \cite{dong2023aligndiff} first demonstrated their effectiveness in this domain. Subsequent works, such as Lu et al. \cite{lu2025makes}, further investigated Transformer-based architectures, showing that long-range attention mechanisms can improve sample quality and enhance model expressiveness.

\paragraph{Training Inefficiency Caused by Slow Convergence in DiTs} 
However, a major challenge remains: Diffusion Transformers typically suffer from slow convergence, which significantly increases training costs. Efforts to improve DiT training efficiency can be broadly categorized into two lines of research: architecture-level modifications and non-model design approaches. In this work, we focus on the latter, which includes training and optimization techniques without altering the model architecture. Notable non-model design approaches include the use of different noise schedules \cite{kingma2023understanding, nichol2021improved}, additional prediction targets such as velocity direction \cite{yao2024fasterdit}, and novel training sampling strategies \cite{esser2024scaling, hang2023efficient}. SiT \cite{ma2024sit}, which incorporates Flow Matching into the diffusion process, also falls into this category, improving training efficiency without changing model structure.

\paragraph{Loss Weighting as a Key Factor in Diffusion Model Convergence}
One particularly relevant direction in non-model design is loss weighting, which significantly influences convergence behavior and sample quality \cite{esser2024scaling, hang2023efficient, karras2024analyzing}. EDM2 \cite{karras2024analyzing} adopts an uncertainty-based multi-task loss to model the varying importance of training signals across noise scales. sCM \cite{lu2024simplifying} further confirms the effectiveness of adaptive weighting, demonstrating that learned weighting functions can outperform heuristic ones. Despite their effectiveness, these approaches often suffer from slow convergence due to the added complexity of learning the weighting function via neural approximators, which can introduce instability during optimization.

\paragraph{Variationally Optimal and Fast-converging Weighting via Online Regression} 
To address these limitations, we propose a new method that derives the variationally optimal form of the weighting function and introduces a closed-form, fast-converging approximation based on online polynomial regression. This method avoids unstable iterative optimization, accelerates training, and improves generalization across noise scales.

% ===== 在main.tex中添加以下内容 =====
\section{Method} \label{sec:method}
This section aims to improve the training efficiency of diffusion planning by designing a principled and uncertainty-aware loss weighting function under the flow-based generative modeling framework. Motivated by the limitations of neural network-based weighting (e.g., MLP), we first introduce a continuous formulation of uncertainty-based loss, then derive the variationally optimal weighting function, followed by a discussion of the MLP-based modeling approach and its drawbacks, and finally present a polynomial-based online estimation method that better approximates the optimal weighting and improves convergence.

\paragraph{(1) Continuous formulation of uncertainty-based loss. }We start by generalizing the uncertainty-based multi-task loss proposed by~\cite{kendall2018multi} to the continuous noise-conditioning setting used in diffusion models.
Let $\sigma \in (0, \frac{\pi}{2})$ denote the continuous noise scale, and let $D_\theta$ be the denoising model with parameters $\theta$. The per-noise loss is denoted as $\mathcal{L}(D_\theta; \sigma)$, and we introduce a log-variance proxy function $u(\sigma)$. The total training objective is reformulated as:

\begin{equation}
    \mathcal{L}_{\text{cont}} = \mathbb{E}_{\sigma \sim p(\sigma)} \left[ \frac{\lambda(\sigma) \cdot \mathcal{L}(D_\theta; \sigma)}{\exp(u(\sigma))} + u(\sigma) \right]
    \label{eq:cont_loss}
\end{equation}

where $\lambda(\sigma)$ is a noise-dependent weighting function.
A detailed derivation of Eq.~\eqref{eq:cont_loss} is provided in Appendix~\ref{app:proof}.

\paragraph{(2) Variationally optimal weighting function. }To find the optimal form of $u(\sigma)$, we treat Equation~\eqref{eq:cont_loss} as a functional and apply variational calculus to minimize it with respect to $u(\cdot)$.

\begin{equation}
    \mathcal{L}[u] = \int \left[ \frac{\lambda(\sigma) \cdot \mathcal{L}(D_\theta; \sigma)}{\exp(u(\sigma))} + u(\sigma) \right] p(\sigma) \, d\sigma
    \label{eq:var_loss}
\end{equation}

Taking the functional derivative and setting it to zero gives:
\begin{equation}
    \frac{\delta \mathcal{L}}{\delta u(\sigma)} = -\frac{\lambda(\sigma) \cdot \mathcal{L}(D_\theta; \sigma)}{\exp(u(\sigma))} + 1 = 0
    \label{eq:var_deriv}
\end{equation}

Solving Equation~\eqref{eq:var_deriv} yields the optimal weighting function:
\begin{equation}
    u^*(\sigma) = \log \lambda(\sigma) + \log \mathcal{L}(D_\theta; \sigma)
    \label{eq:u_star}
\end{equation}

\begin{figure}[t]
    \centering
    \begin{subfigure}[b]{0.3\linewidth}
        \includegraphics[width=\linewidth]{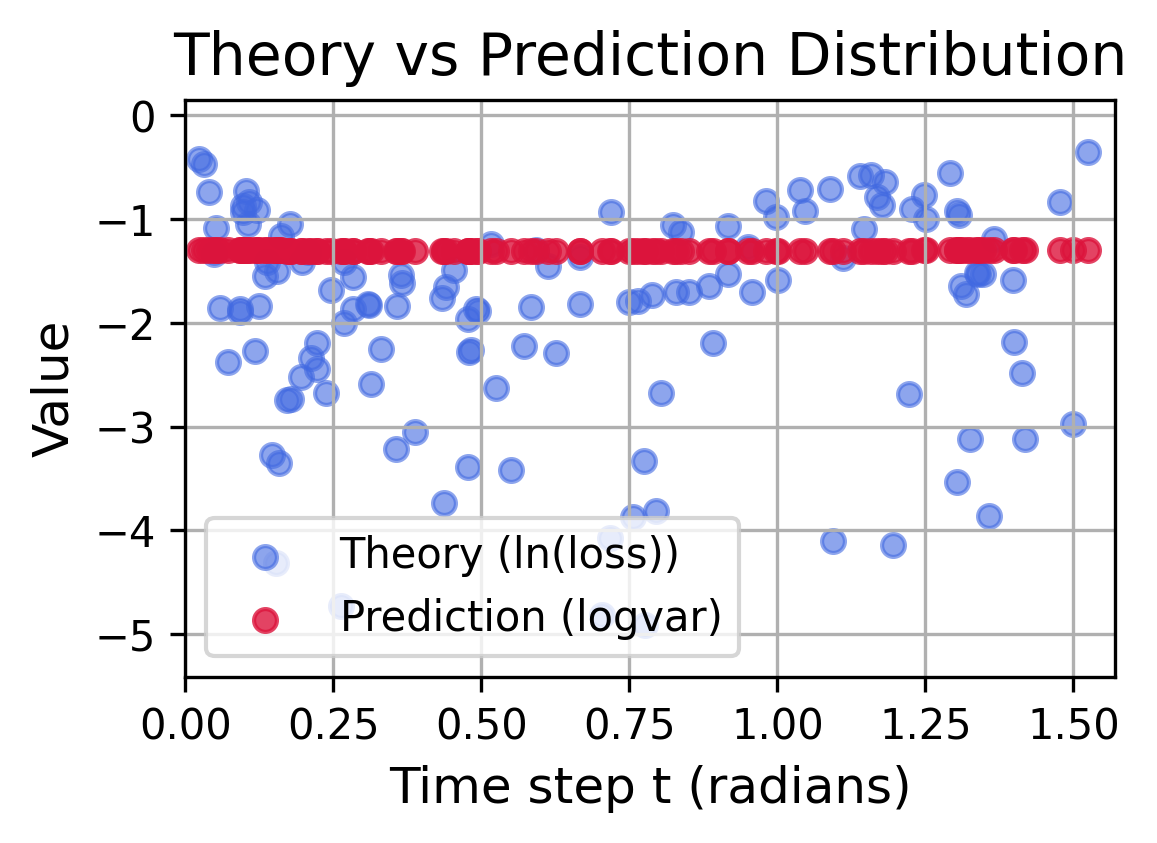}
        \caption*{(a)}
    \end{subfigure}
    \hfill
    \begin{subfigure}[b]{0.3\linewidth}
        \includegraphics[width=\linewidth]{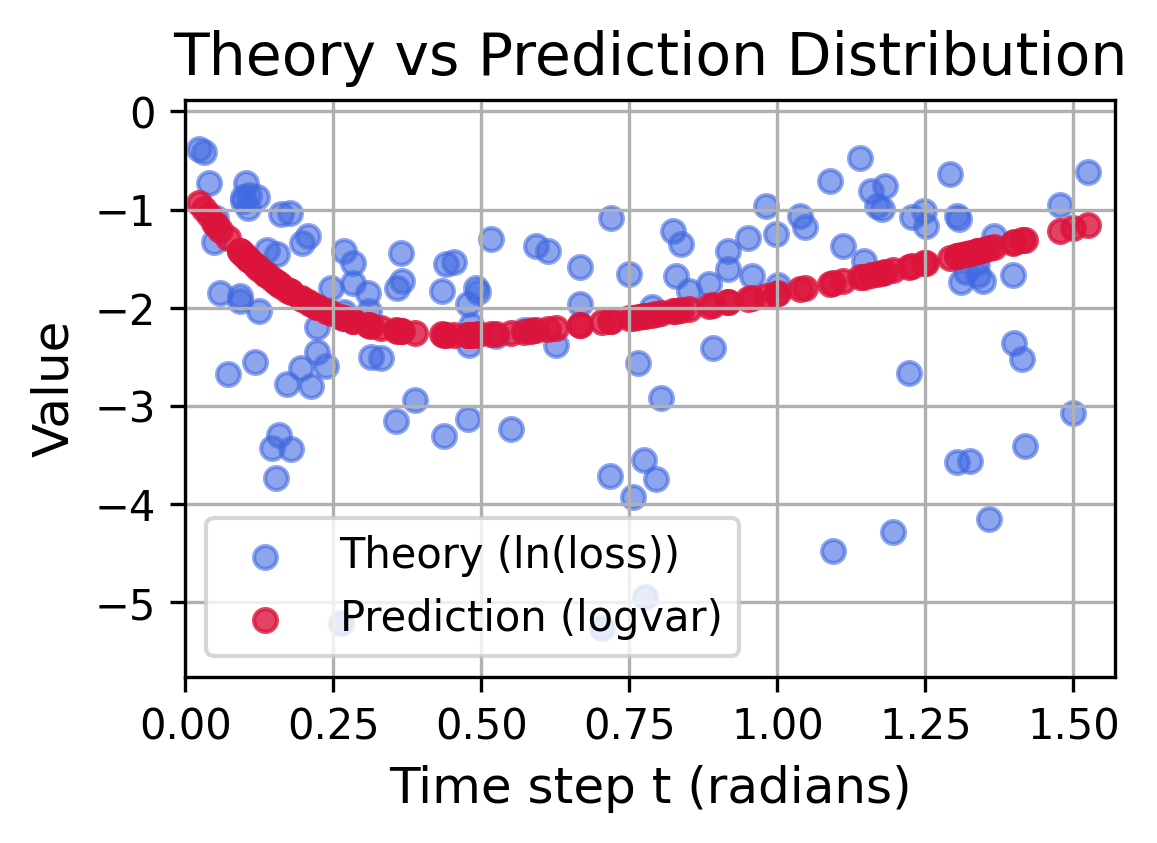}
        \caption*{(b)}
    \end{subfigure}
    \hfill
    \begin{subfigure}[b]{0.32\linewidth}
        \centering
        \includegraphics[width=\linewidth]{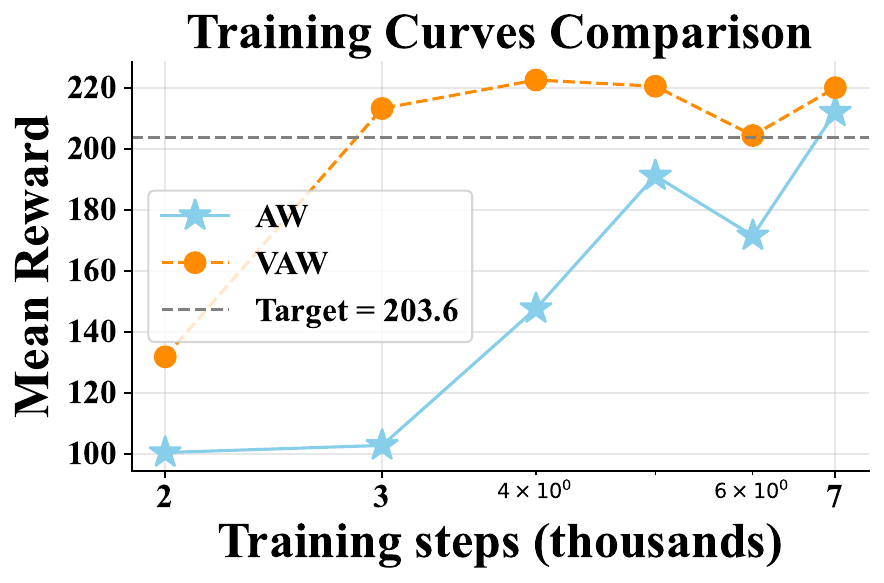}
        \caption*{(c)}
    \end{subfigure}

    \caption{
        \textbf{Comparison between MLP and Polynomial approximation.}
        (a) \& (b): Early-stage prediction of per-noise uncertainty. The MLP (a) remains flat initially, struggling to capture sharp uncertainty structure, while the Polynomial model (b) fits faster and improves convergence.  
        (c): Maze2d-large performance. Polynomial weighting surpasses the baseline within 3k steps; MLP takes around 7k.
    }
    \label{fig:poly-vs-mlp}
\end{figure}

\paragraph{(3) MLP-based modeling of the weighting function. } To instantiate $u(\sigma)$, EDM2\cite{karras2024analyzing} propose a learnable function approximator using a simple neural network:
\begin{equation}
    u(\sigma) \approx \text{MLP}(c_{\text{noise}}(\sigma))
    \label{eq:mlp_approx}
\end{equation}
where $c_{\text{noise}}(\sigma)$ denotes the Fourier-feature encoding of $\sigma$. The MLP is trained jointly with $D_\theta$ and discarded at test time.

Although conceptually simple, this approach has notable limitations. The MLP struggles to track the sharp variations of $\mathcal{L}(D_\theta; \sigma)$, especially during early training, where data is sparse and noisy. As a result, the predicted $u(\sigma)$ often collapses to a nearly constant value. This can lead to poor weighting across noise levels, as illustrated qualitatively in Figure~\ref{fig:poly-vs-mlp}, where the estimated uncertainty distribution exhibits minimal structure.

\paragraph{(4) Polynomial approximation for online estimation. }
The optimal weighting function $u^*(\sigma)$ in Eq.~\eqref{eq:u_star} depends directly on $\log \mathcal{L}(D_\theta; \sigma)$, which may vary significantly over training. Rather than relying on a parameter-heavy neural approximator such as an MLP, we propose a direct polynomial regression approach to estimate this log-loss curve efficiently.

Let $x_i := \log \sigma_i$ and $y_i := \log \mathcal{L}(D_\theta; \sigma_i)$ for a set of sampled noise levels $\{\sigma_i\}$ and their corresponding training losses. We fit a polynomial of degree $d$,
\begin{equation}
    P(x) = \sum_{k=0}^{d} w_k x^k,
    \label{eq:poly}
\end{equation}
such that $P(\log \sigma) \approx \log \mathcal{L}(D_\theta; \sigma)$. The overall approximation of the optimal uncertainty weight then becomes:
\begin{equation}
    u(\sigma) = \log \lambda(\sigma) + P(\log \sigma).
    \label{eq:u_poly}
\end{equation}

The polynomial coefficients $w_k$ are computed using least-squares fitting over a mini-batch of $(\log \sigma_i, \log \mathcal{L}_i)$ pairs. To ensure temporal smoothness during training, we maintain an exponential moving average of the estimated coefficients across batches. This yields a fully online, closed-form estimator that does not rely on backpropagation or learned network parameters.

\begin{algorithm}[H]
    \caption{Streaming Polynomial Fitting for Uncertainty Weighting}
    \label{alg:PolyFit}
    \textbf{Input:} Degree $d$, EMA coefficient $\mu$, batch $\{(\sigma_i, \mathcal{L}_i)\}_{i=1}^B$ \\
    \textbf{Output:} Uncertainty function $u(\sigma)$ \\
    Transform pairs: $x_i \gets \log \sigma_i,\; y_i \gets \log \mathcal{L}_i$ \\
    Construct Vandermonde matrix: $X_{ij} \gets x_i^j$ for $0 \leq j \leq d$ \\
    Compute polynomial coefficients: $\hat{w} \gets (X^\top X)^{-1} X^\top y$ \\
    Update weights: $w \gets \mu w + (1 - \mu) \hat{w}$ \\
    Output: $u(\sigma) \gets \log \lambda(\sigma) + \sum_{k=0}^d w_k (\log \sigma)^k$
\end{algorithm}

\paragraph{Advantages.}
Compared to the MLP-based modeling (see Eq.~\eqref{eq:mlp_approx}), this polynomial approximation approach offers the following practical benefits:
\begin{itemize}
    \item \textbf{Fast-converging:} The model adapts rapidly to new loss trends across $\sigma$, even in the early training phase.
    \item \textbf{Non-iterative:} Closed-form regression avoids the overhead of gradient-based updates.
    \item \textbf{Closed-form:} All computations are analytic and lightweight, suitable for online updates.
\end{itemize}

As illustrated in Figure~\ref{fig:poly-vs-mlp}, our polynomial scheme captures the structure of the per-noise loss more faithfully than MLP-based modeling, especially when the true uncertainty landscape is sharply non-uniform. This results in faster convergence and improved weighting accuracy across noise scales.

% --- Experiments Section ---
\section{Experiments} \label{sec:exp}

The primary objective of our experiments is to evaluate the training efficiency of our proposed method---a flow-based generative model with variational adaptive weighting---when used for trajectory modeling in planning tasks. Rather than aiming to surpass state-of-the-art (SOTA) performance in absolute terms, our evaluation focuses on comparing convergence speed. To this end, we propose Flow Veteran (FV), a variant of the Diffusion Veteran (DV) framework where its trajectory modeling component is systematically replaced with our method. This allows us to assess improvements in training efficiency.

\subsection{Experimental Setup}

\paragraph{Testing via Module Replacement.}
To isolate and evaluate the impact of our trajectory modeling strategy, we modify the Diffusion Veteran (DV) framework by replacing its planner's trajectory generation module---originally based on the VP-SDE formulation---with our flow matching-based model. Apart from this substitution, the rest of the DV architecture remains unchanged, ensuring a fair and consistent comparison. We choose DV due to its rigorous empirical design and its status as a SOTA method that captures key insights from prior diffusion planning studies. Implementation and training details for the integration are further described in Appendix~\ref{app:details}.

\paragraph{Choice of Flow Model.}
Among various flow matching approaches, we adopt \textbf{Trigflow} as the default trajectory generator throughout our experiments. This decision is supported by empirical evidence: as shown in Section~\ref{app:ablations}, Trigflow consistently outperforms linear flow matching on a range of tasks in terms of convergence speed. Additionally, recent works such as sCM~\citep{lu2024simplifying} have successfully employed Trigflow in large-scale image generation tasks, further demonstrating its robustness and practical value.

\paragraph{Benchmarks}
We conducted experiments on the D4RL dataset~\citep{fu2020d4rl}, one of the most widely used benchmarks for offline RL. The dataset covers diverse domains, enabling evaluation across navigation, and manipulation tasks. We selected the following tasks to cover different trajectory planning challenges:

\begin{itemize}
    \item \textbf{Maze2D}: A 2D maze navigation task, assessing an agent's spatial planning ability in environments with geometric constraints.
    \item \textbf{Franka Kitchen}: A high-dimensional robotic manipulation task with multi-stage goals, designed to test both long-horizon reasoning and trajectory precision.
\end{itemize}

\paragraph{Implementation Details}
All experiments were conducted on  a single computing platform equipped with NVIDIA GeForce RTX 4090 GPUs. Additional training configurations are provided in Appendix~\ref{app:details}.

\subsection{Results and Q\&As}

\input{resources/tables/simple_table}
\paragraph{Q1: Does FV perform better across benchmarks?}

\textbf{A:} \textit{Yes. FV achieves the best performance on all benchmark tasks, including both locomotion (Maze2D) and manipulation (Kitchen) domains.}

Table~\ref{table:overall-performance} presents a comprehensive performance comparison across tasks. Our method outperforms all previous approaches.
We compare our method with a range of baselines commonly used in offline RL and behavior planning literature. The compared methods fall into three categories:
\begin{itemize}
    \item \textbf{Gaussian policies}: including BC: vanilla imitation learning, CQL~\citep{kumar2020conservative}, IQL~\citep{kostrikov2021offline};
    \item \textbf{Diffusion policies}: including SfBC~\citep{chen2022offline}, DQL~\citep{wang2022diffusion}, IDQL~\citep{hansen2023idql};
    \item \textbf{Diffusion planners}: including Diffuser~\citep{janner2022planning}, Adaptdiffuser (AD)~\citep{liang2023adaptdiffuser}, Decision Diffusion (DD)~\citep{ajay2022conditional}, Hierachical Diffuser (HD)~\citep{chen2024simple}, and Diffusion Veteran (DV)~\citep{dong2024cleandiffuser}.
\end{itemize}

The results of the above methods are obtained from their respective literature, except for DV*, which we replicate following the cleandiffuser codebase~\citep{dong2024cleandiffuser}. Notably, our focus is not on outperforming these methods, but rather on validating the training efficiency of our proposed trajectory modeling approach under a consistent planning framework.

Our approach reaches state-of-the-art results with significantly fewer training steps: only \textbf{3k} steps on Maze2D tasks and \textbf{40k} steps on Kitchen tasks, compared to \textbf{1M} steps commonly used in prior works. Furthermore, our model uses only \textbf{5 inference steps} compared to \textbf{20 steps} in VP-SDE, demonstrating efficiency. These results highlight the scalability and generalization of our approach across diverse settings while maintaining low computation budgets.

\begin{figure*}[h]
    \centering
    \begin{subfigure}[b]{0.32\linewidth}
        \includegraphics[width=\linewidth]{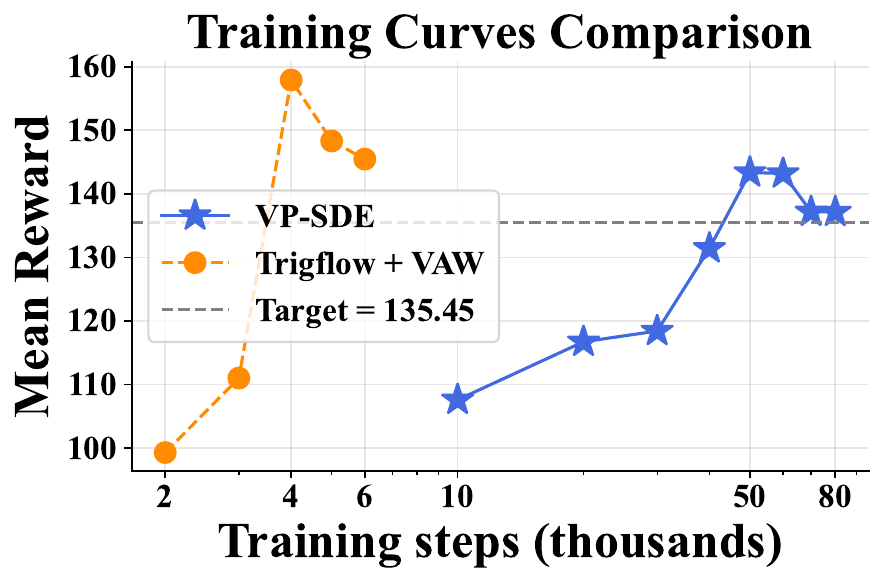}
        \caption{}
        \label{fig:maze2d-umaze}
    \end{subfigure}
    \hfill
    \begin{subfigure}[b]{0.32\linewidth}
        \includegraphics[width=\linewidth]{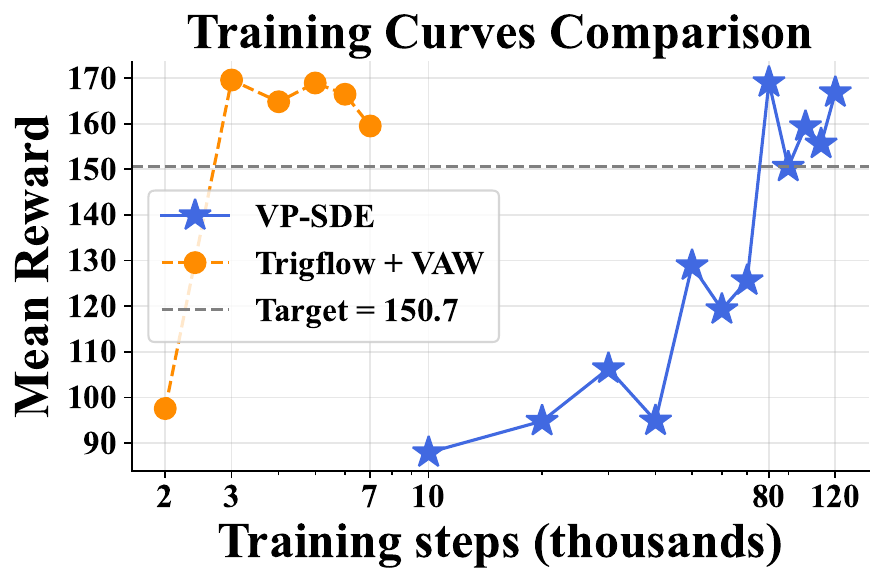}
        \caption{}
        \label{fig:maze2d-medium}
    \end{subfigure}
    \hfill
    \begin{subfigure}[b]{0.32\linewidth}
        \includegraphics[width=\linewidth]{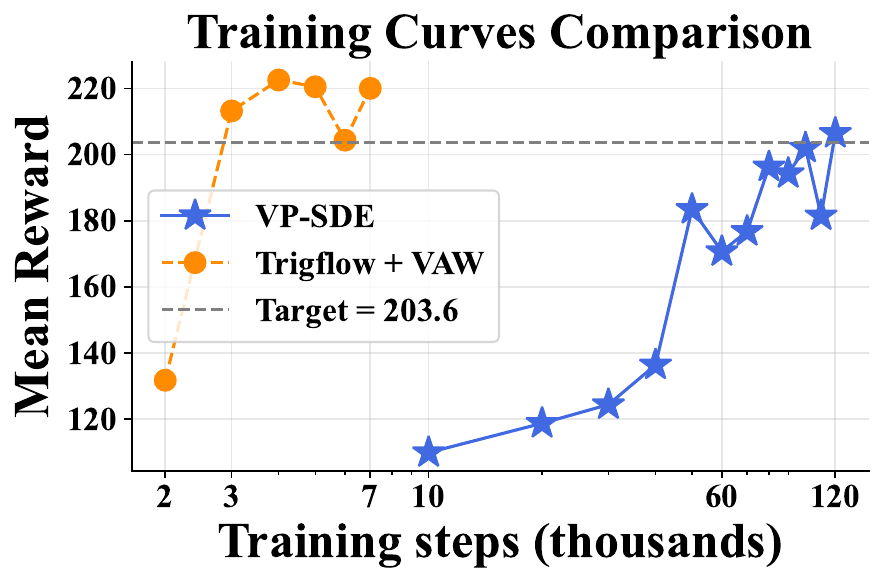}
        \caption{}
        \label{fig:maze2d-large}
    \end{subfigure}

    \begin{center}
        \begin{subfigure}[b]{0.32\linewidth}
            \includegraphics[width=\linewidth]{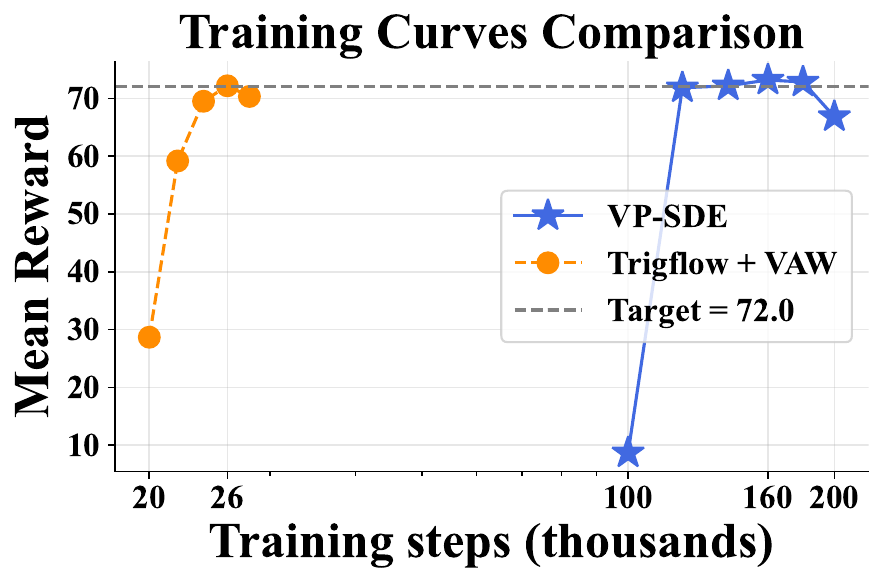}
            \caption{}
            \label{fig:kitchen-mixed}
        \end{subfigure}
        \hspace{0.06\linewidth}
        \begin{subfigure}[b]{0.32\linewidth}
            \includegraphics[width=\linewidth]{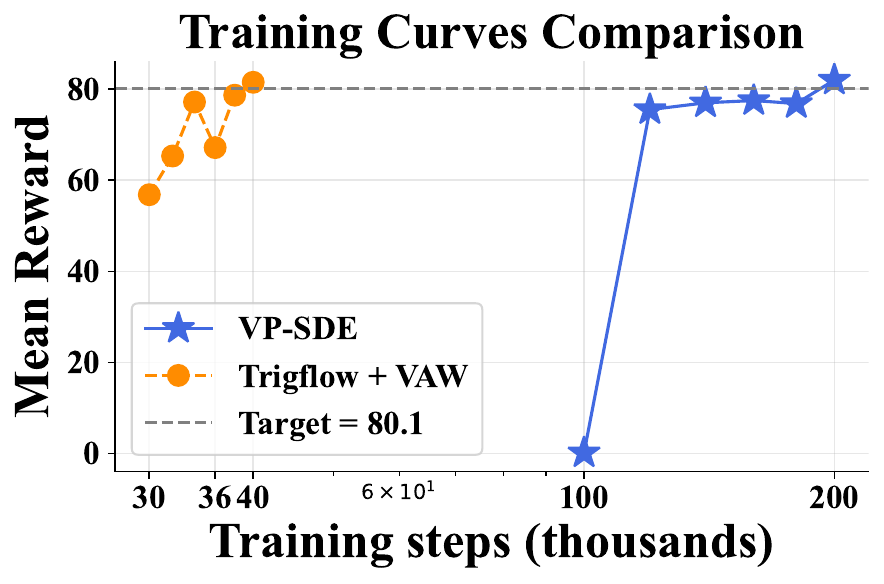}
            \caption{}
            \label{fig:kitchen-partial}
        \end{subfigure}
    \end{center}

    \caption{
        \textbf{TrigFlow+VAE (FV) vs. VP-SDE in Different Tasks.}
        (a)-(c): Maze2D navigation tasks (U-Maze, Medium, Large);  
        (d)-(e): Kitchen manipulation tasks (Mixed, Partial).  
        \textbf{FV} (our method) consistently reaches state-of-the-art performance with substantially fewer training steps: from \textbf{5×} speedup in Kitchen tasks to \textbf{12×–40×} in Maze2D tasks. This demonstrates superior convergence.
    }
    \label{fig:convergence_all}
\end{figure*}

\paragraph{Q2: What is the convergence speed of FV compared to VP-SDE?}

\textbf{A:} \textit{As shown in Figure~\ref{fig:convergence_all}, FV achieves state-of-the-art performance with drastically fewer training steps.}

In the Maze2D locomotion benchmarks, FV consistently reaches or surpasses the prior state-of-the-art with drastically fewer training steps. On the maze2d-umaze task, FV exceeds the previous SOTA at just 3k training steps, whereas VP-SDE requires 120k steps to achieve comparable performance. On the maze2d-medium task, FV again surpasses SOTA in 3k steps, compared to 80k steps for VP-SDE. In the more complex maze2d-large task, FV reaches SOTA at 4k steps, while VP-SDE requires 50k.

In Kitchen manipulation tasks, FV maintains its efficiency advantage. On the kitchen-partial task, FV surpasses the previous SOTA at 40k steps, whereas VP-SDE needs 200k steps to reach the same baseline. Similarly, in the kitchen-mixed task, FV reaches SOTA at 26k steps, while VP-SDE requires 140k steps.

These results translate into a convergence speedup of up to $40\times$ in Maze2D locomotion tasks and around $5\times$ in complex kitchen manipulation tasks, clearly illustrating the superior sample efficiency and rapid convergence of FV across domains.

\paragraph{Q3: Which Weighting Function performs better?}

\begin{figure}[htbp]
    \centering
    \begin{subfigure}[t]{0.31\textwidth}
        \centering
        \includegraphics[width=\textwidth]{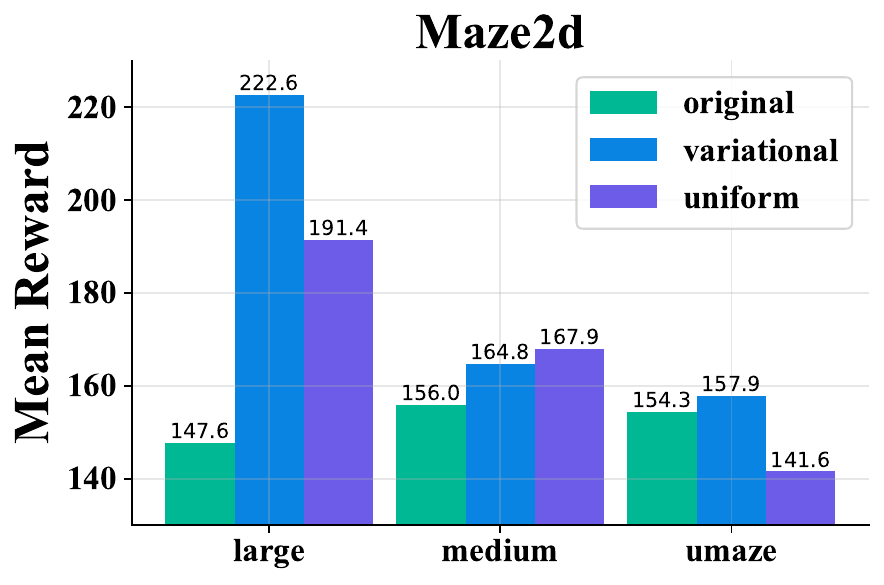}
        \caption{}
        \label{fig:maze2d_ablation}
    \end{subfigure}
    \hfill
    \begin{subfigure}[t]{0.31\textwidth}
        \centering
        \includegraphics[width=\textwidth]{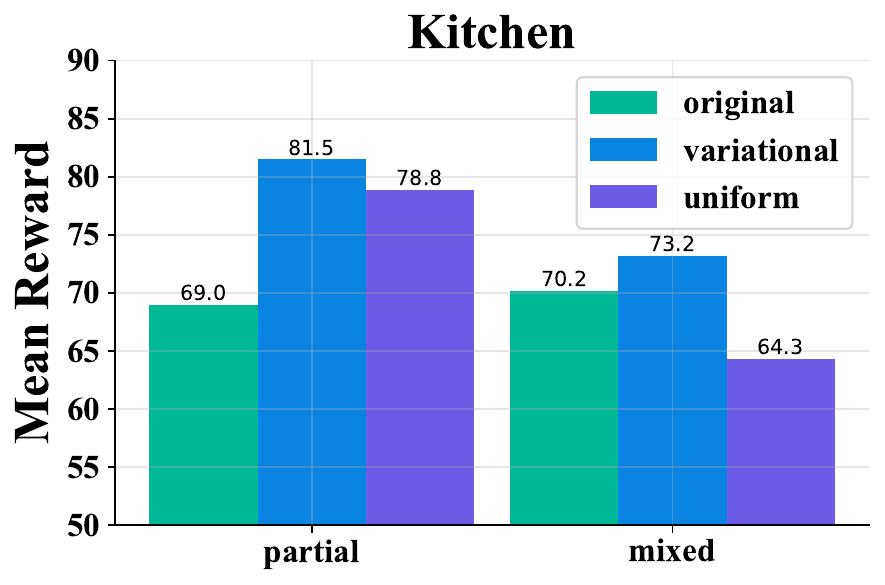}
        \caption{}
        \label{fig:kitchen_ablation}
    \end{subfigure}
    \hfill
    \begin{subfigure}[t]{0.31\textwidth}
        \centering
        \includegraphics[width=\textwidth]{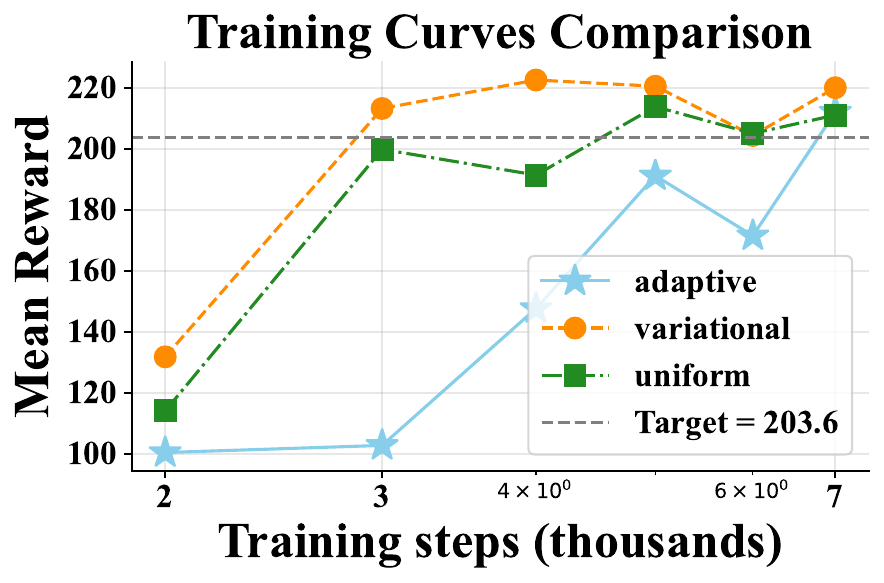}
        \caption{}
        \label{fig:maze2d_large_performance}
    \end{subfigure}
        \caption{Comparison of weighting functions across tasks.  
        (a) Maze2D performance at 4k steps.  
        (b) Kitchen performance at 40k steps.  
        (c) Training curves on Maze2D-Large, where variational adaptive weighting (labeled as \textit{variational}) achieves the fastest convergence.  
        Original adaptive weighting and uniform weighting are labeled as \textit{origin} and \textit{uniform}, respectively.}
    \label{fig:trigflow_ablation}
\end{figure}

\textbf{A:} The variational adaptive weighting method converges the fastest across most tasks, particularly excelling in more challenging environments.

As illustrated in Figure~\ref{fig:trigflow_ablation}, variational adaptive weighting consistently demonstrates superior convergence behavior compared to uniform weighting and original adaptive weighting (MLP-based). In all three Maze2D environments—\textit{umaze}, \textit{medium}, and \textit{large}—the variant with variational adaptive weighting rapidly achieves higher mean rewards with significantly fewer training steps. This is especially evident in the most difficult setting, maze2d-large, where variational adaptive weighting exhibits a steep learning curve early in training, reaching near-optimal performance well before the other two methods begin to stabilize.

%% file: resources/tables/simple_table.tex
\begin{table*}[t]
\centering
\small
\captionsetup{font=small, labelfont=bf}
\caption{\textbf{Performance of various offline-RL methods.} Our method achieves the best performance across all benchmark tasks. Our results are averaged over 150 episode seeds. We omit the variance over seeds for simplicity. The best average performance on each task set is marked in bold fonts.}
\label{table:overall-performance}
\resizebox{\textwidth}{!}{
\begin{tabular}{llrrrrrrrrrrrr}
\toprule
\textbf{Dataset} & \textbf{Env} & \textbf{BC} & \textbf{CQL} & \textbf{IQL} & \textbf{SfBC} & \textbf{DQL} & \textbf{IDQL} & \textbf{Diffuser} & \textbf{AD} & \textbf{DD} & \textbf{HD} & \textbf{DV*} & \textbf{FV (Ours)} \\
\midrule
Kitchen & Mixed & 47.5 & 51.0 & 51.0 & 45.4 & 62.6 & 66.5 & 52.5 & 51.8 & 75.0 & 71.7 & 72.0 & 72.5 \\
        & Partial & 33.8 & 49.8 & 46.3 & 47.9 & 60.5 & 66.7 & 55.7 & 55.5 & 56.5 & 73.3 & 80.1 & 82.1 \\
        & avg. & 40.7 & 50.4 & 48.7 & 46.7 & 61.6 & 66.6 & 54.1 & 53.7 & 65.8 & 72.5 & 76.1 & \textbf{77.3} \\
\midrule
Maze2D  & Large & 0.0 & 57.5 & 43.6 & 74.4 & -- & 90.1 & 123.0 & 167.9 & -- & 128.4 & 203.0 & 222.6 \\
        & Medium & 0.0 & 15.4 & 70.6 & 73.8 & -- & 89.5 & 121.5 & 129.9 & -- & 135.6 & 150.7 & 164.8 \\
        & Umaze & 0.0 & 36.4 & 57.1 & 73.9 & -- & 57.9 & 113.9 & 135.1 & -- & 155.8 & 134.5 & 157.9 \\
        & avg. & 5.0 & 12.5 & 58.6 & 74.0 & -- & 79.2 & 119.5 & 144.3 & -- & 139.9 & 163.6 & \textbf{181.8} \\
\bottomrule
\end{tabular}
}
\end{table*}

%% file: resources/appendix.tex
\section{Ablation Study} \label{app:ablations}

\begin{figure}[h]
    \centering
    \begin{subfigure}[b]{0.32\linewidth}
        \includegraphics[width=\linewidth]{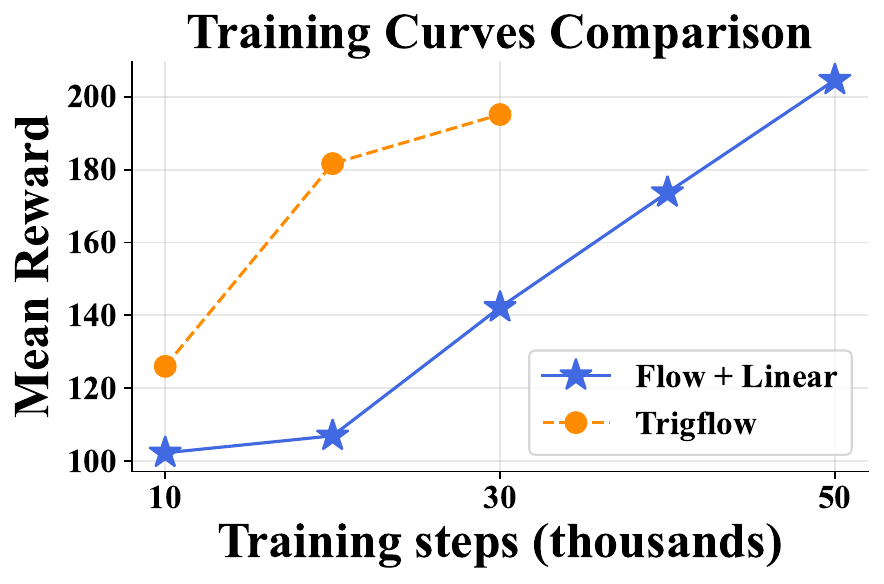}
        \caption*{(a)}
    \end{subfigure}
    \hfill
    \begin{subfigure}[b]{0.32\linewidth}
        \includegraphics[width=\linewidth]{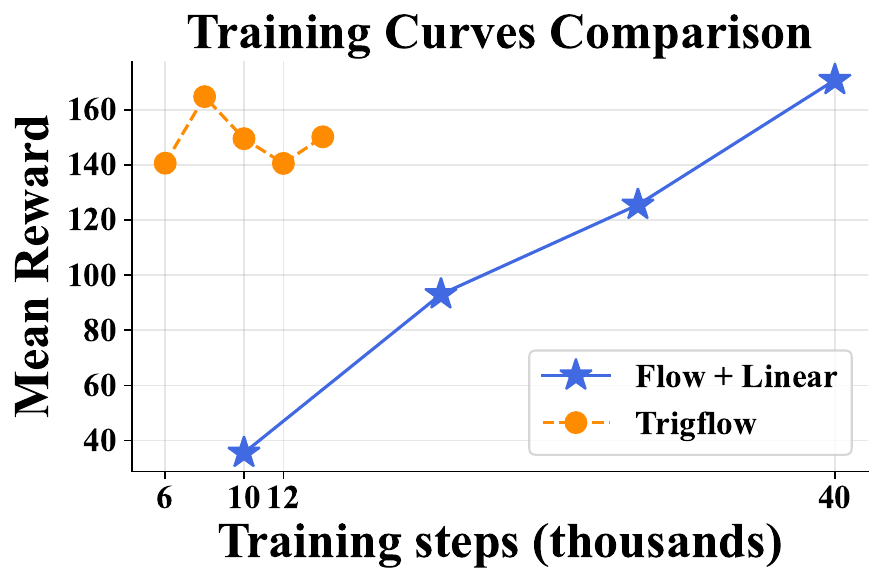}
        \caption*{(b)}
    \end{subfigure}
    \hfill
    \begin{subfigure}[b]{0.32\linewidth}
        \centering
        \includegraphics[width=\linewidth]{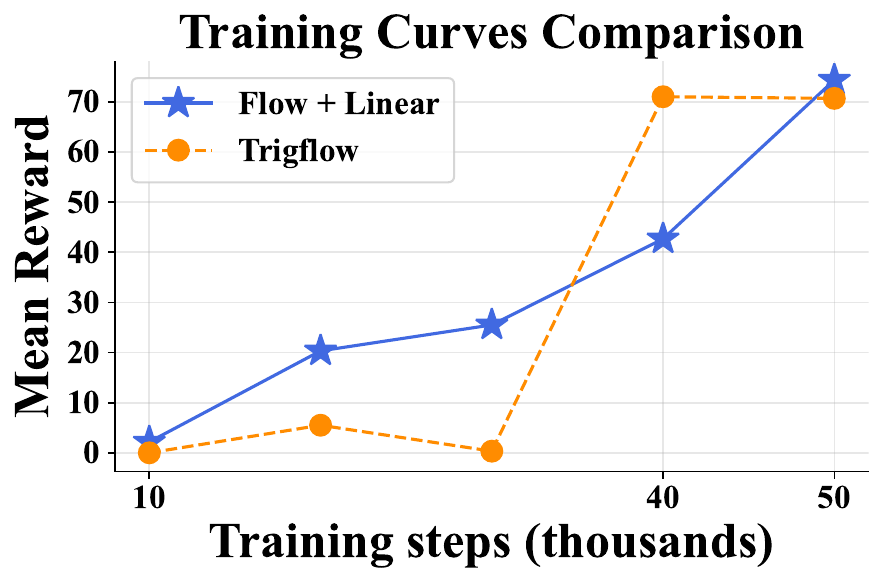}
        \caption*{(c)}
    \end{subfigure}
    
    \caption{Ablation studies comparing TrigFlow and Flow Matching with a linear path and uniform time distribution. (a) In Maze2D-Large, TrigFlow reaches similar performance in fewer steps. (b) In Maze2D-Medium, TrigFlow converges significantly faster. (c) In Kitchen-Mixed, TrigFlow again reaches target performance in fewer iterations.}
    \label{fig:flow_ablation_figures}
\end{figure}

We perform an ablation study to assess the impact of using TrigFlow, which incorporates a trigonometric time transformation and a non-linear path, compared with the standard Flow Matching method using a linear path and uniform time distribution. This analysis is designed to examine whether TrigFlow's design choices contribute to practical improvements, particularly in convergence speed.

\paragraph{Experimental setup}
To isolate the effect of the path and time distributions, we keep all other training and inference conditions consistent across the two methods. Both models use uniform weighting during training and a fixed learning rate of $8 \times 10^{-4}$. The same number of inference steps (10) is used, and the exponential moving average (EMA) rate is set to 0.9999. TrigFlow adopts DDIM for sampling, whereas Flow Matching uses Euler integration. The time distributions are defined as uniform over $[0,1]$ for Flow Matching and logit-normal with parameters $(-0.4, 1.6)$ for TrigFlow.

\paragraph{Results}
We evaluate both methods on three control tasks: Maze2D-Large, Maze2D-Medium, and Kitchen-Mixed. In each task, TrigFlow demonstrates a consistently faster convergence under the same budget. In Maze2D-Large, TrigFlow reaches competitive performance at 30k steps, while Flow Matching requires 50k. In Maze2D-Medium, the performance difference is more prominent, with convergence observed at 8k versus 40k steps. Kitchen-Mixed shows similar results, with TrigFlow achieving stable performance at 40k steps, compared to 50k for Flow Matching. These findings suggest that the proposed path and time designs can reduce training cost while maintaining performance.

\section{Implementation and Training Details} \label{app:details}

\paragraph{Planner Implementation.} 
We train only the \textbf{planner module}, where the Flow Models adopt the same network backbone as DV's original planner. Specifically, we use DiT1D~\cite{peebles2023scalable, dong2023aligndiff}, with a hidden dimension of 256, head dimension of 32, and 2 transformer blocks. This setup is consistent across all tasks.

\paragraph{Training Details.}
All models are optimized using Adam~\cite{kingma2014adam}, with a learning rate of $8 \times 10^{-4}$ for Flow Models and a batch size of 128. The number of gradient steps is 4k for Maze2D and 40k for Kitchen tasks. For diffusion training, we use the following settings: $P_{\text{mean}} = -0.4$, $P_{\text{std}} = 1.6$, and $\sigma_{\text{data}} = 1$. We use 5 sampling steps with the DPM-Solver-2M sampler. The exponential moving average (EMA) rates are set to $[0.999, 0.9995]$, and the polynomial order used in TrigFlow is 5.

\paragraph{Auxiliary Components: Diffusion Veteran (DV).}
Apart from the planner, all other components in our method directly follow \textbf{Diffusion Veteran (DV)}. These include the trajectory critic $V(x(\tau))$, which estimates expected returns from predicted state trajectories and is used to bias the planner toward high-return behaviors, and the inverse dynamics model $\epsilon_\omega$, which recovers the corresponding action sequences from predicted state sequences. We follow DV's implementation and training setup without modification. The full design, including these auxiliary modules, is detailed in DV~\cite{lu2025makes}, which provides a strong empirical foundation and has been widely adopted.

\section{Theoretical Analysis}
\label{app:proof}

\paragraph{Continuous Formulation of Uncertainty-based Loss.}

We derive the continuous form of the uncertainty-weighted loss used in the main text (Eq.~\eqref{eq:cont_loss}) by generalizing the discrete formulation from~\cite{kendall2018multi}.

\subparagraph{Discrete Uncertainty-weighted Loss.}
In multi-task learning, given a set of task-specific losses $\{\mathcal{L}_i\}_{i=1}^N$, the uncertainty-based formulation assigns each loss a learned variance $\sigma_i^2 > 0$, yielding:
\begin{equation}
\mathcal{L}_{\text{discrete}} = \sum_{i=1}^N \left( \frac{\mathcal{L}_i}{2\sigma_i^2} + \frac{1}{2}\log \sigma_i^2 \right).
\label{eq:app_discrete}
\end{equation}

Rewriting with $u_i := \log \sigma_i^2$ and ignoring the constant $1/2$ (since it does not affect optimization), we obtain:
\begin{equation}
\mathcal{L}_{\text{discrete}} = \sum_{i=1}^N \left( \mathcal{L}_i e^{u_i} + u_i \right).
\label{eq:app_discrete_log}
\end{equation}

\subparagraph{Generalization to Continuous Noise Conditioning.}
In the diffusion setting, each noise scale $\sigma \in (0, \frac{\pi}{2})$ can be viewed as defining a separate task.
Let $\mathcal{L}(D_\theta; \sigma)$ denote the loss at noise scale $\sigma$, and define a continuous log-variance function $u(\sigma)$ in place of the discrete set $\{u_i\}$.
To reflect the distribution over $\sigma$, we introduce a sampling distribution $p(\sigma)$ and a weighting function $\lambda(\sigma)$ that aligns with prior work (e.g., EDM2~\cite{karras2024analyzing}).

The continuous counterpart to Eq.~\eqref{eq:app_discrete_log} becomes:
\begin{equation}
\mathcal{L}_{\text{cont}}(D_\theta, u) = \int_0^{\frac{\pi}{2}} \lambda(\sigma)\left( \mathcal{L}(D_\theta; \sigma) e^{u(\sigma)} + u(\sigma) \right) p(\sigma)\, d\sigma.
\label{eq:app_continuous_integral}
\end{equation}

Or, equivalently, in expectation form:
\begin{equation}
\mathcal{L}_{\text{cont}}(D_\theta, u) = \mathbb{E}_{\sigma \sim p(\sigma)} \left[ \lambda(\sigma) \left( \mathcal{L}(D_\theta; \sigma) e^{u(\sigma)} + u(\sigma) \right) \right].
\label{eq:app_continuous_expectation}
\end{equation}

This matches Eq.~\eqref{eq:cont_loss} in the main text. The formulation reflects a continuous generalization of the discrete uncertainty-weighted loss, where the task index is replaced by a continuously sampled noise scale $\sigma$ and the log-variance becomes a learnable function $u(\sigma)$.

\qed